\documentclass[final,5p,times,twocolumn]{elsarticle}

\usepackage{amssymb}
\usepackage{psfrag,epsfig,subfigure}
\usepackage{amsmath}
\usepackage{url}
\usepackage[utf8]{inputenc} 
\usepackage[T1]{fontenc}    
\usepackage{hyperref}       
\usepackage{url}            
\usepackage{booktabs}       
\usepackage{amsfonts}       
\usepackage{nicefrac}       
\usepackage{microtype}      
\usepackage{lipsum}
\usepackage{fancyhdr}       
\usepackage{graphicx}       
\graphicspath{{media/}}     
\usepackage{psfrag,epsfig,subfigure}

\journal{Computers in Biology and Medicine}

\begin{document}

\begin{frontmatter}



\title{Dual-Task Vision Transformer for Rapid and Accurate Intracerebral Hemorrhage CT Image Classification}


\author[label1]{Jialiang Fan}
\author[label2,label3]{Xinhui Fan\corref{cor2}}
\author[label2]{Chengyan Song}
\author[label3]{Xiaofan Wang} 
\author[label3]{Bingdong Feng} 
\author[label4]{Lucan Li}
\author[label1,label5,label6]{Guoyu Lu}

\affiliation[label1]{organization={Franklin College of Arts and Sciences, University of Georgia},
            city={Athens},
            state={GA},
            country={USA}}

\affiliation[label2]{organization={Department of Neurology, Yulin Hospital, The First Affiliated Hospital of Xi'an Jiaotong University},
            city={Yulin},
            country={China}}

\affiliation[label3]{organization={Department of Neurology, The First Hospital of Yulin},
            city={Yulin},
            country={China}}

\affiliation[label4]{organization={School of Art, Lanzhou University},
            city={Lanzhou},
            country={China}}

\affiliation[label5]{organization={College of Agricultural \& Environmental Sciences, University of Georgia},
            city={Athens},
            state={GA},
            country={USA}}

\affiliation[label6]{organization={College of Engineering, University of Georgia},
            city={Athens},
            state={GA},
            country={USA}}

\cortext[cor2]{Corresponding author. \textit{Email address:} fanxinhui89@foxmail.com (X. Fan).}

\begin{abstract}
Intracerebral hemorrhage (ICH) is a severe and sudden medical condition caused by the rupture of blood vessels in the brain, leading to permanent damage to brain tissue and often resulting in functional disabilities or death in patients. Diagnosis and analysis of ICH typically rely on brain CT imaging. Given the urgency of ICH conditions, early treatment is crucial, necessitating rapid analysis of CT images to formulate tailored treatment plans. However, the complexity of ICH CT images and the frequent scarcity of specialist radiologists pose significant challenges. Therefore, we collect a dataset from the real world for ICH and normal classification and three types of ICH image classification based on the hemorrhage location, i.e., Deep, Subcortical, and Lobar. In addition, we propose a neural network structure, dual-task vision transformer (DTViT), for the automated classification and diagnosis of ICH images. The DTViT deploys the encoder from the Vision Transformer (ViT), employing attention mechanisms for feature extraction from CT images. The proposed DTViT framework also  incorporates two multilayer perception (MLP)-based decoders to simultaneously identify the presence of ICH and classify the three types of hemorrhage locations. Experimental results demonstrate that DTViT performs well on the real-world test dataset. The code and newly collected dataset for this work are available at: \url{https://github.com/jfan1997/DTViT}.

\end{abstract}

\begin{keyword}
Intracerebral hemorrhage, image classification, vision transformer, transfer learning.
\end{keyword}

\end{frontmatter}



\section{Introduction}\label{sec1}
Intracerebral hemorrhage (ICH) is a type of severe condition characterized by the formation of a hematoma within the brain parenchyma \cite{gebel2000intracerebral,de2020surgery}. Representing 10-15\% of all stroke cases, ICH is linked to significant morbidity and mortality rates \cite{wan2023brain}. Head computerized tomography (CT) is the standard method to diagnose ICH that can obtain accurate images of the head anatomical structure and detect abnormalities. However, analyses of head CT images for ICH classification and diagnosis usually require skilled radiologists, who are in very short supply in developing countries or regions. This may lead to delays in setting appropriate treatment plans and interventions, which are very crucial for accurate ICH patients. 

Furthermore, the diagnosis and analysis of ICH in CT images often require extensive experience and concentrated attention. However, when physicians are overworked or face a high volume of cases, misdiagnoses can inevitably occur. Therefore, the use of assistive medical technologies to aid physicians in diagnosing and analyzing ICH images can enhance the speed of diagnosis and treatment. This not only alleviates the shortage of medical resources but also improves diagnostic efficiency and accuracy, which are of significant importance to both physicians and patients.

In recent years, computer vision methods based on deep learning have played a crucial role in CT imaging diagnostics. Neural network models trained on extensive real-world data have demonstrated high accuracy and efficiency on medical images\cite{kim2022transfer,jiang2023review,butoi2023universeg,yuan2023effective}. There are studies that have utilized computer vision and deep learning techniques for ICH image detection and classification to aid medical professionals and enhance diagnostic efficiency \cite{cortes2023deep,chen2022deep}. In  \cite{cortes2023deep}, a deep learning model based on ResNet and EfficientDet that detects bleeding in CT scans is proposed, offering both classification and region-specific decision insights and achieving an accuracy of 92.7\%. Literature \cite{chen2022deep} demonstrates the efficacy of convolutional neural network (CNN)-based deep learning models, particularly CNN-2 and ResNet-50,  in classifying strokes from CT images, with future plans to optimize these models for improving diagnostic accuracy and efficiency. Li {\it et al.} \cite{li2020deep} propose a Unet-based neural network model to detect hemorrhage strokes of CT images, achieving an accuracy of 98.59\%.

Datasets are essential for training high-performance neural network models, particularly in the context of medical imaging. However, challenges such as limited access to diverse cultural artifacts, varying image quality standards, and restrictions on image usage rights add complexity to the creation of high-quality ICH image datasets. Several publicly available ICH image datasets exist, such as the brain CT images with intracranial hemorrhage masks published on Kaggle, which includes 2,500 CT images from 82 patients, though it is relatively small in size \cite{vbookshelf_computed_2024}. Another dataset contains high-resolution brain CT images with 2,192 sets of images for segmentation \cite{wu2023bhsd}. Additionally, the RSNA 2019 public dataset, with 874,035 CT images, represents the largest collection available \cite{flanders2020construction}.

Compared to existing datasets, our dataset offers unique characteristics. Firstly, our data was collected over the past four years, from 2020 to 2024, allowing it to reflect the symptom characteristics of ICH under recent social conditions. Secondly, our dataset includes 15,936 CT images from 249 patients, with images meticulously filtered by medical physicians to remove meaningless noise. Finally, and most importantly, our ICH CT images are labeled based on hemorrhage location, which is critical for diagnosis and treatment planning for ICH patients.


Transformer, a network architecture based on the self-attention mechanism, has achieved remarkable success in the field of natural language processing (NLP) in recent years \cite{vaswani2017attention}. Building on this success, researchers have extended Transformer to the field of computer vision, introducing Vision Transformer (ViT) \cite{dosovitskiy2020image}. The ViT model segments images into patches and feeds them into the Transformer, and then,  computes attention weights between different pixel blocks, facilitating effective feature extraction from images. Experimental results demonstrate that Vision Transformers achieve superior and more promising performance compared to classical CNN-based neural networks. 

Transfer learning is widely used in deep learning to reduce training time and improve training efficiencies. A pre-trained model is firstly built on a large benchmark dataset such as ImageNet \cite{deng2009imagenet}, and common features are captured and saved in the weights of models. When applying the model to a new dataset or different tasks, we can train the model using task-specific datasets on top of the pre-trained model. Transfer learning also reduces the need for large labeled datasets, making deep learning on small or imbalanced datasets possible, which is particularly suitable for medical images as medical images are usually difficult to obtain.

Building on the aforementioned discussions, we first build an image dataset from real-world sources that includes head images of both healthy individuals and ICH patients, categorized into three types—Deep, Lobar, and Subtentorial—based on the location of the hemorrhage. Furthermore, a dual-task Vision Transformer (DTViT)  is designed to simultaneously classify images of normal individuals and ICH patients, as well as categorize three types of ICH based on the hemorrhage location. We conduct experiments using the new dataset and the proposed model, and the results show that our proposed model achieves superior testing accuracy of $99.88\%$. The contributions of this paper are summarized as follows:
\begin{itemize}
 \item We have constructed a real-world ICH image dataset comprising CT images of both normal individuals and patients classified into three types of ICH, addressing issues of insufficient brain hemorrhage image datasets and the lack of datasets for brain hemorrhage location, which is of great significance to both medical and computer vision research.
\item Built upon the constructed dataset, we have developed a deep learning model, i.e., dual-task Vision Transformer (DTViT), which is capable of performing dual-classification tasks simultaneously: distinguishing CT images of normal and ICH patients and identifying the type of ICH according to the hemorrhage location. 
\item Our comparative experiments show that the DTViT model achieves 99.88\% accuracy, outperforming existing CNN models and demonstrating the high quality of our dataset.
\end{itemize}
The remainder of this paper is structured as follows: Section \ref{s2} describes the dataset and data preprocessing techniques. Section \ref{s3} introduces architectures of ViT, DTViT, and transfer learning for image classification and diagnosis. Section \ref{s4} details the experimental setup, evaluation metrics, and the results obtained. Section \ref{s5} discusses the limitations of this study and directions for future research. Finally, Section \ref{s6} concludes the paper.

\section{Data Collection and Image Processing Methodology}\label{s2}
In this section, we introduce the newly collected dataset, the data preprocessing of ICH images, and the construction of DTVIT for image diagnosis.
\begin{figure}[h!]
  \centering
   \includegraphics[scale=0.65]{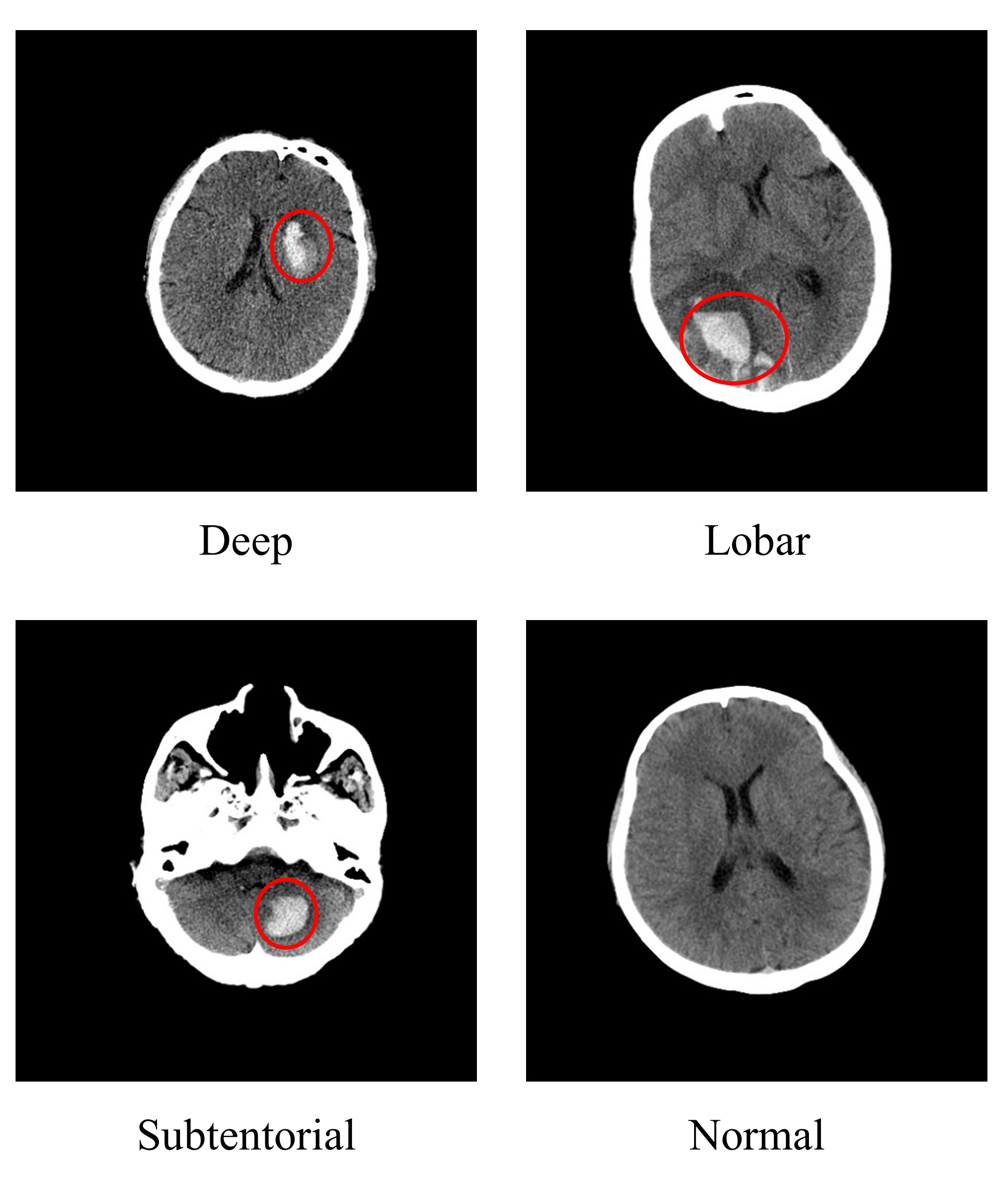} 
  \caption{Normal and brain hemorrhages in three different locations.}
  \label{fig:four_types}
\end{figure}
\subsection{Dataset}\label{s2.1}
The dataset is sourced from the Department of Neurology at The First Hospital of Yulin. It includes 15,936 CT slices from 249 patients with intracerebral hemorrhage (ICH) collected between 2018 and 2021, and 6,445 CT slices from 199 healthy individuals in 2024. The healthy subjects have one set of CT images, while the ICH patients have two sets, captured within 24 hours and within 72 hours of symptom onset separately. All images were obtained using the GE LightSpeed VCT scanner at Yulin First Hospital.
All scans are saved as DCM files, featuring a resolution of 512×512 pixels, a slice thickness of 5 mm, and an inter-slice gap of either 5 mm or 1 mm.

In addition, each group of ICH images is manually classified into three different types according to hemorrhage location by an expert physician: Deep ICH, Lobar ICH, and Subtentorial ICH.
Figure \ref{fig:four_types} shows a sample image that includes three types of hemorrhages along with normal images.

To protect patient's privacy, the dataset only includes patients' gender and CT images. A detailed data description is listed in Table \ref{tab:table1}.
\renewcommand{\arraystretch}{1.3}
\begin{table}[h!]
 \caption{Data distribution of the dataset.}
  \centering
  \begin{tabular}{lllll}
    \toprule
    Attribute & Total    & Value     & Number & Percentage  \\
    \midrule
    \textbf{Sex} & {$221$} & Male  & $117$ & $52.94\%$ \\ 
    & &Female   &  $104$ &$47.06\%$ \\

    \textbf{ICH} & $12651$ & Yes  & $8244$ & $65.16\%$ \\ 
    & &No  &  $4407$ &$34.84\%$ \\
    \textbf{Location} & 8244 & Deep  & $6093$  & $73.91\%$    \\
  & & Lobar  & $1656$ & $20.09\%$     \\
    & & Subtentorial  & $495$ & $6\%$    \\
    \bottomrule
  \end{tabular}
  \label{tab:table1}
\end{table}

\subsection{Data preprocessing}
The data preprocessing process is composed of three stages: morphological treatment, manual filtering, and data augmentation.

\subsubsection{Morphological treatment}
When taking CT images, patients are equipped with fixation braces to keep their heads steady, which is also captured by the scanner and may add noise to the image data. Therefore, it is necessary to remove the artifact to purify the data.
We apply morphological processing to remove the fixation brace and normalize the images, converting them from DICOM to JPG format. The steps are as follows.

Initially, the pixel array is extracted from the DICOM file and duplicated twice.  The first duplicate is used to create a mask, while the second is used to generate the final output. We start by binarizing the pixel array to emphasize key features and apply erosion using a disk-shaped element to minimize noise. Next, the edge columns are zeroed out to remove potential artifacts, and flood filling is performed to enhance specific areas, thus producing the mask image.

Subsequently, this mask is applied to the second, normalized pixel array. The resulting image is then converted to 8-bit integers and saved as a JPG file. An instance of the original CT image and the processed result is displayed in Fig. \ref{fig:fig1}. It can be seen that the fixed brace has been successfully removed and the normalized image is much clearer than the original.

\subsubsection{Manual filtering}

\begin{figure}[h!]
  \centering
   \includegraphics[scale=0.7]{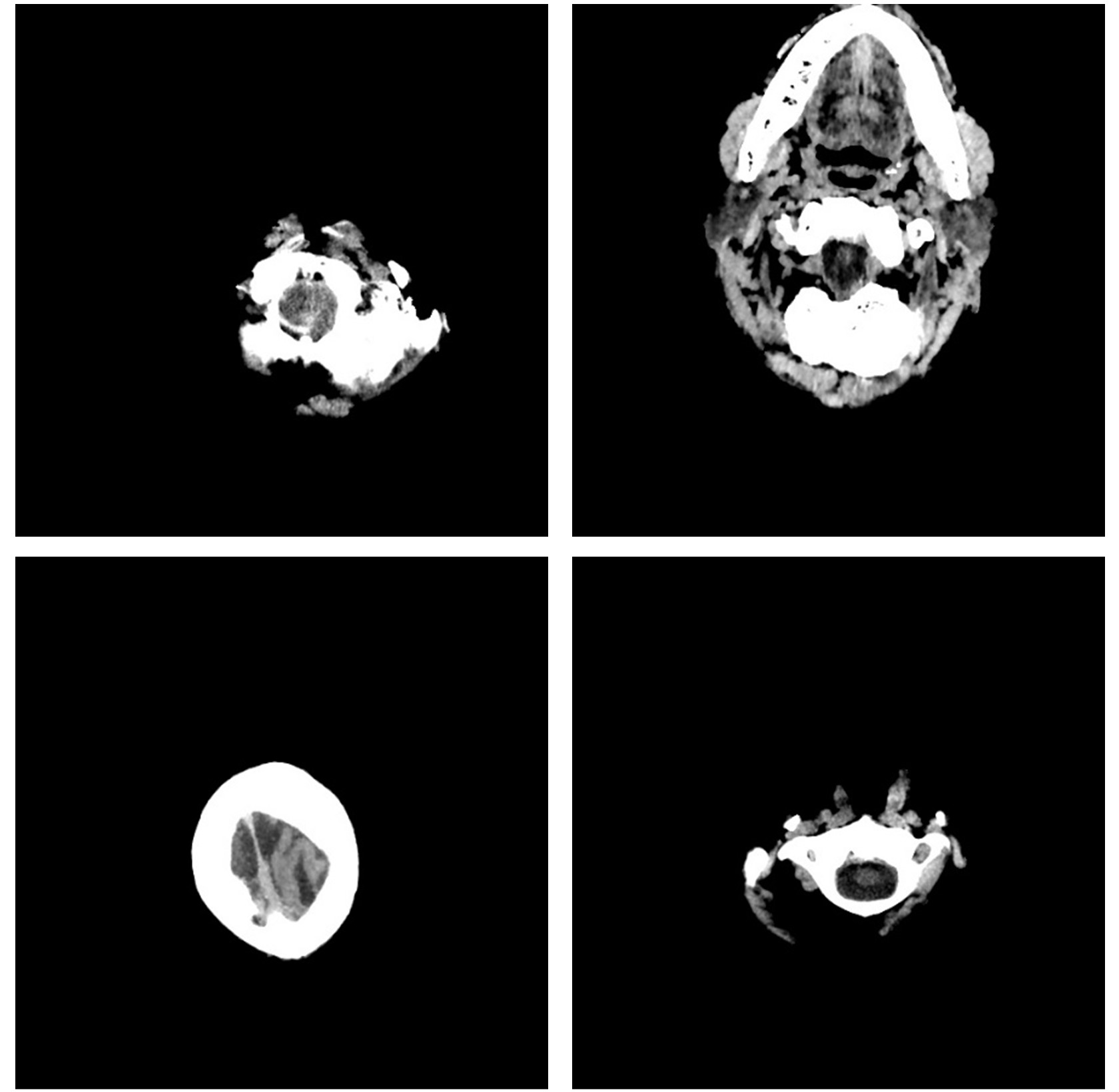} 
  \caption{Noise images that have been removed from the dataset.}
  \label{fig:removed}
\end{figure}

The head CT scan usually starts from the base of the brain (near the neck) and covers the entire brain up to the forehead. This means that only part of CT scans can capture the hemorrhage location and present a clear view for diagnoses, while other CT images do not include effective information and are useless for classification. Therefore, medical specialists manually diagnose and select each group of CT images to remove meaningless images, refine the dataset,  and reduce noise. This step is critical to constructing the dataset. Figure \ref{fig:removed} shows sample images that were removed from the dataset. These images do not exhibit any signs of hemorrhage.

\begin{figure*}
  \centering
   \includegraphics[scale=0.45]{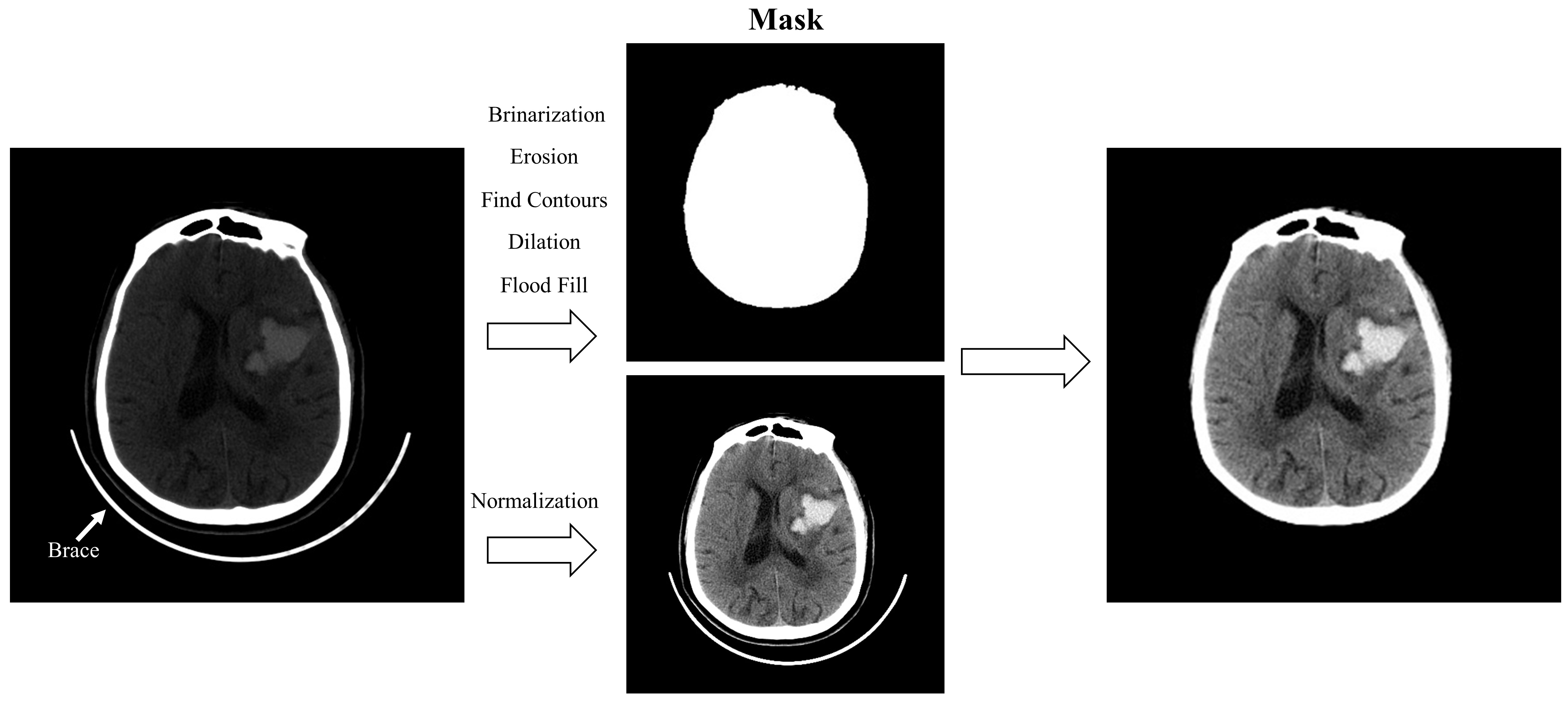} 
  \caption{Morphological treatment of CT images.}
  \label{fig:fig1}
\end{figure*}
\subsubsection{Data augumentation} 
 As presented in Table 1, The number of images for three types of cerebral hemorrhage locations varies significantly. Therefore, we perform data augmentation operations on subtentorial CT images. Firstly, we augmented the dataset of three types of cerebral hemorrhage CT images proportionally to ensure that the distribution was approximately 1:1:1. Secondly, we also augmented the images without cerebral hemorrhage to achieve a roughly 1:1 ratio between the two categories. The augmented dataset consists of 30,222 images in total, including 13,221 normal images and 17,001 ICH images, which are further categorized into 6,093 Deep, 5,940 Lobar, and 4,968 Subtentorial images. Furthermore, to enhance the diversity of the dataset, we applied image transformation on the training dataset, including center cropping to the size of 224x224, random rotation of 15 degrees, random sharpness adjustment with a factor of 2, and normalization with the mean value $[0.485,0.456,0.406]$ and standard deviation  $[0.229,0.224,0.225]$.



 \begin{figure*}
  \centering
\includegraphics[scale=0.2]{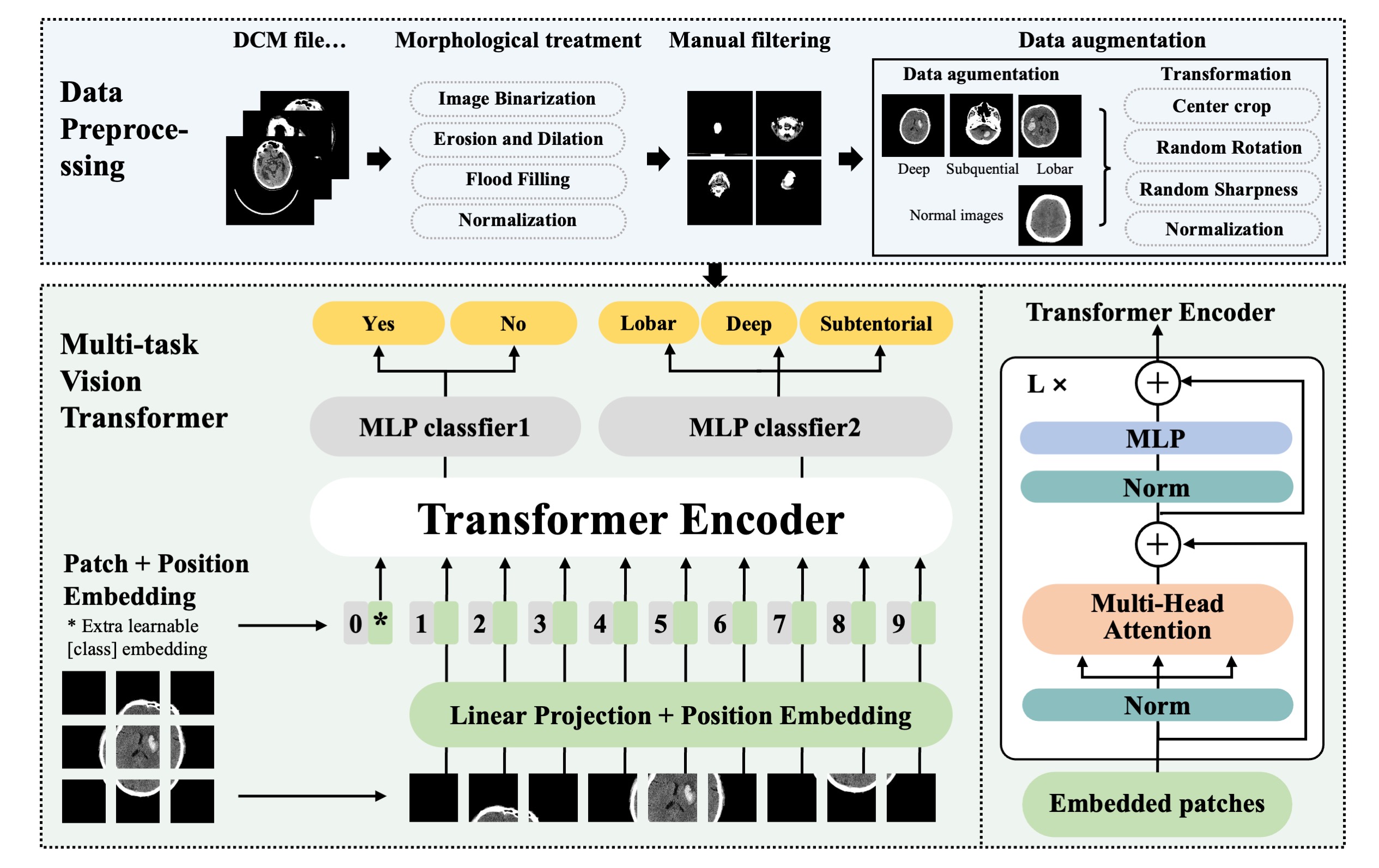}
  \caption{The research diagram of the DTViT model.}
  \label{fig:fig2}
\end{figure*}

\section{DTViT for ICH CT Image Diagnosis}\label{s3}
This section introduces the structure of DTViT. We first introduce the
architecture of DTViT and then discuss the transfer learning and details of the pre-trained model. An overall diagram is shown in Fig. \ref{fig:fig2}.

\subsection{Structure of Vision Transformer}
Vision Transformer (ViT) \cite{dosovitskiy2020image} firstly applies the self-attention mechanism to image classification tasks, which has obtained excellent performances. The ViT, as shown in Fig. \ref{fig:fig2}, consists of several key components: patch embedding, position embedding, Transformer encoder, classification head, and other optional components. The processing procedure is as follows. Firstly,  a raw 2D image is initially transformed into a sequence of 1D patch embeddings, which mirrors the word embedding technique used in natural language processing. Similarly, the positional embeddings are applied to the constructed patch embeddings to retain information about the location of each patch in the original image. Then, the sequences are fed into the encoder of the Transformer, which consists of multi-head self-attention layers and MLP layers, with normalization and residual connections. It enables the model to capture intricate connections among different patch blocks. Finally, the output is processed by the decoder, i.e., the full-connection layer or MLP layer, to output the classification results.

The details of crucial procedures are as follows.  When an $c\times h \times w$ 2D image ($c$ represents the  channel number, $h$ represents image heights, and $w$ represents image width) is fed into the model. The first step is patching embedding, where the image is cut into $N$ image pieces, where $N=hw/p^{2}$, and $p$ is the specified patch size of the image piece. Then, these image pieces are flattened as a vector with size of  $p^{2}c$ and  linear projected to a lower-dimensional space as 
\begin{equation}
    z_{i} =W\cdot x_{i}+b,
\end{equation}where $W\in\mathbb{R}^{D\times p^{2}c}$ is the weight matrix; $x_{i}\in \mathbb{R}^{p^{2}c}$ represents the flattened image vector, and $z_{i}\in \mathbb{R}^{D}$ represents the image vector after projection. Then,  a token $z_{class}\in \mathbb{R}^{D}$ for classification and an embedding recording of the positional information are applied to the vector, yielding the final inputs to the Transformer as 
\begin{equation}\nonumber
   input= [z_{class}; z_{i}+pos_{1};z_{i}+pos_{2};...;z_{i}+pos_{n}],
\end{equation}where $pos_{i}\in\mathbb{R}^{D}$ represent positional embeddings to retain spatial information of image pieces.

Then, the input is processed by the Transformer's encoder. As shown in Fig. \ref{fig:fig2}, the encoder consists of \( L \) identical blocks, each containing normalization layers, a multi-head attention (MHA) layer, and an MLP layer. The normalization layer is applied before and after the MHA layer, represented as
\begin{equation}
 \text{LN}(x) = \frac{x - \mu}{\sqrt{\sigma^{2} + \epsilon}},
\end{equation}
where \(\mu\) represents the mean of the vector, \(\sigma\) represents the variance of the vector, and \(\epsilon\) is a small constant for numerical stability. Subsequently, the MHA operation is applied to the vector as
\begin{subequations}\label{velocity-kinematic}
     \begin{eqnarray}
&&  Q = XW_{Q},~K = XW_{K},~V = XW_{V}, \\
&& \text{head}_{i} = \text{Attention}(QW_{Q}^{i}, KW_{K}^{i}, VW_{V}^{i}), \\
&& \text{MHA}(Q,K,V) = \text{Concat}(\text{head}_1, \ldots, \text{head}_i)W_{O}, 
 \end{eqnarray}
\end{subequations}
where \( X \) is the input after the layer normalization; \( W_{Q} \), \( W_{K} \),  \( W_{O} \), and \( W_{V} \) are learnable weight matrices; \( Q \), \( K \), and \( V \) represent the calculated queries, keys, and values.
Additionally, the attention calculation formula is as follows:
\begin{equation}
    \text{Attention}(Q,K,V) = \text{softmax}\left(\frac{QK^{\text{T}}}{\sqrt{d}}\right)V,
\end{equation}where \( d \) is the scaling factor, which is the dimension of the query, key, and value vectors. Finally, the MHA output is added to the initial input, followed by a layer normalization, and then fed into the MLP layer to produce the output. This process is repeated iteratively for 
$L$ layers to obtain the final output.

\subsection{Structure of DTViT}
Given the high-density shadows, variable scales, and diverse locations characteristic of ICH CT images, we have adapted the ViT encoder for use in the DTViT. Additionally, we have integrated two MLP-based decoders into the DTViT to facilitate dual-task classification. As illustrated in Fig. \ref{fig:fig2}, both decoders utilize the feature extraction results from the encoder. The first MLP classifier determines whether a hemorrhage is present in the image, while the second MLP classifier identifies the type of hemorrhage based on the location information extracted by the encoder. Correspondingly, the loss function is defined as 
\begin{equation}
    loss=0.5*loss_1+0.5*loss_2,
\end{equation}where $loss_1$ represents the loss of classifier 1 and $loss_2$ represents the loss of classifier 2. This dual-classification approach leverages shared features to enhance the accuracy and efficiency of hemorrhage detection and classification.

\subsection{Transfer learning}
Transfer learning refers to training a model that has already been pre-trained on another large dataset, such as ImageNet \cite{deng2009imagenet}. This approach can significantly save time and data size requirements, and enhance training efficiency. It allows the model to leverage previously learned features and knowledge, which is especially useful when dealing with similar but new tasks. Therefore, we train the DTViT based on a pre-trained Vision Transformer encoder \cite{rw2019timm} as the backbone and fine-tune the model's parameters. 

Specifically, there are three types of pre-trained ViT models: ViT-Base, ViT-Large, and ViT-Huge. The ViT-Base model, with its 12 blocks, is suitable for datasets of moderate complexity. The ViT-Large model, which has 24 blocks and larger embedding dimensions, offers better performance but requires more computational resources. The ViT-Huge is the largest one of the three models, taking 32 blocks and being suitable for high-complexity tasks. For our work with DTViT, we use the ViT-Large as the pre-trained model \cite{touvron2021training}, which is trained on ImageNet-1K \cite{deng2009imagenet}. The configuration of DTViT is shown in Table \ref{tab:ViT-Large}.

\setlength{\tabcolsep}{20pt} 
\renewcommand{\arraystretch}{1.3} 

\begin{table}[h!]
 \caption{Configuration and Parameters of the DTViT.}
 \centering
 \begin{tabular}{cc}
    \toprule
    \textbf{Configuration} & \textbf{Value} \\
    \midrule
    Patch Size & 16 \\
    Embedding Dimension & 1024 \\  
    Attention Heads & 16 \\
    MLP Dimension & 4096 \\
    Parameters & 304326632 \\
    \bottomrule
 \end{tabular}
 \label{tab:ViT-Large}
\end{table}

\begin{figure*}[ht]
  \centering
     \subfigure[]{\includegraphics[scale=0.5]{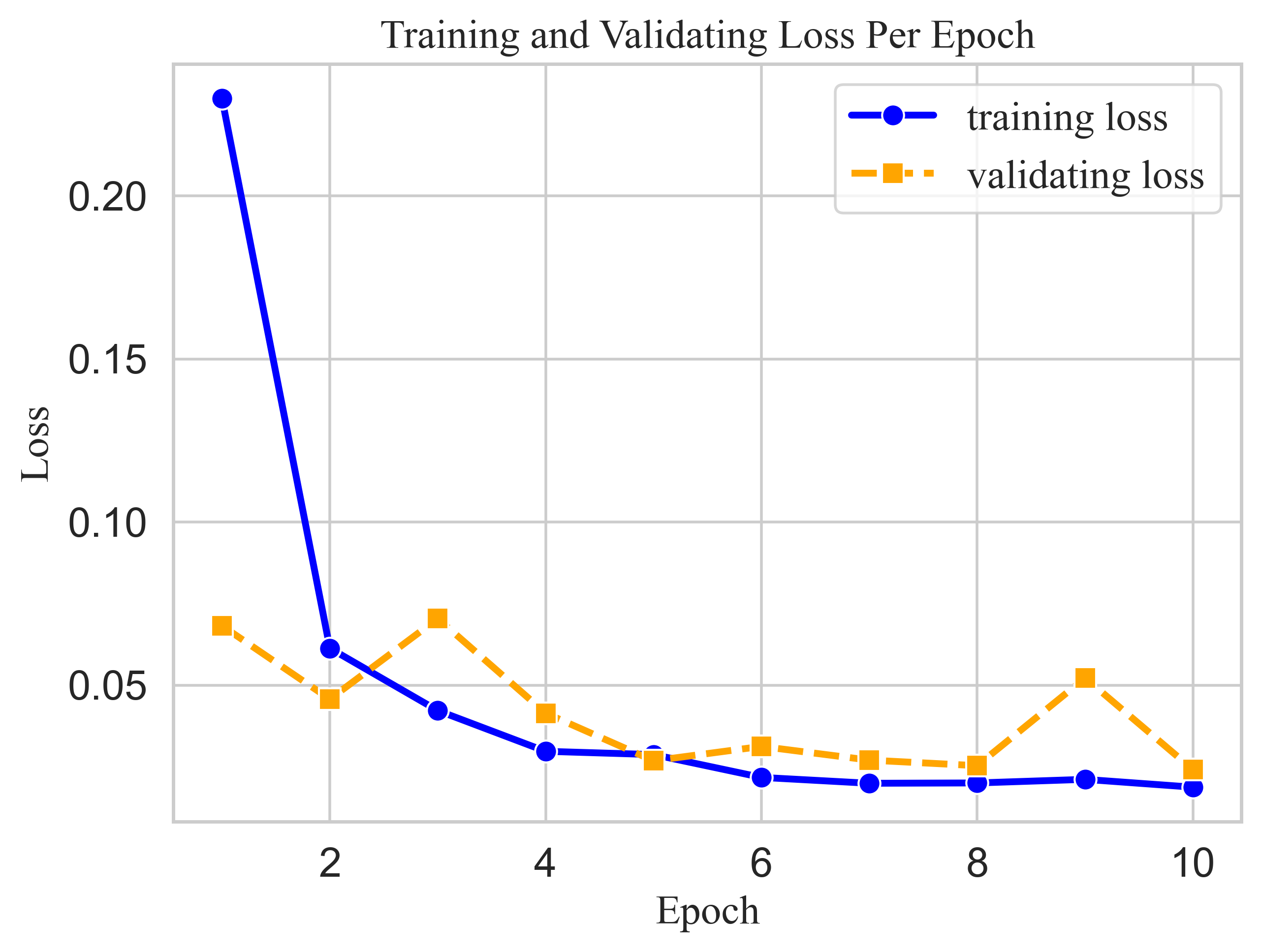}}
 \subfigure[]{\includegraphics[scale=0.5]{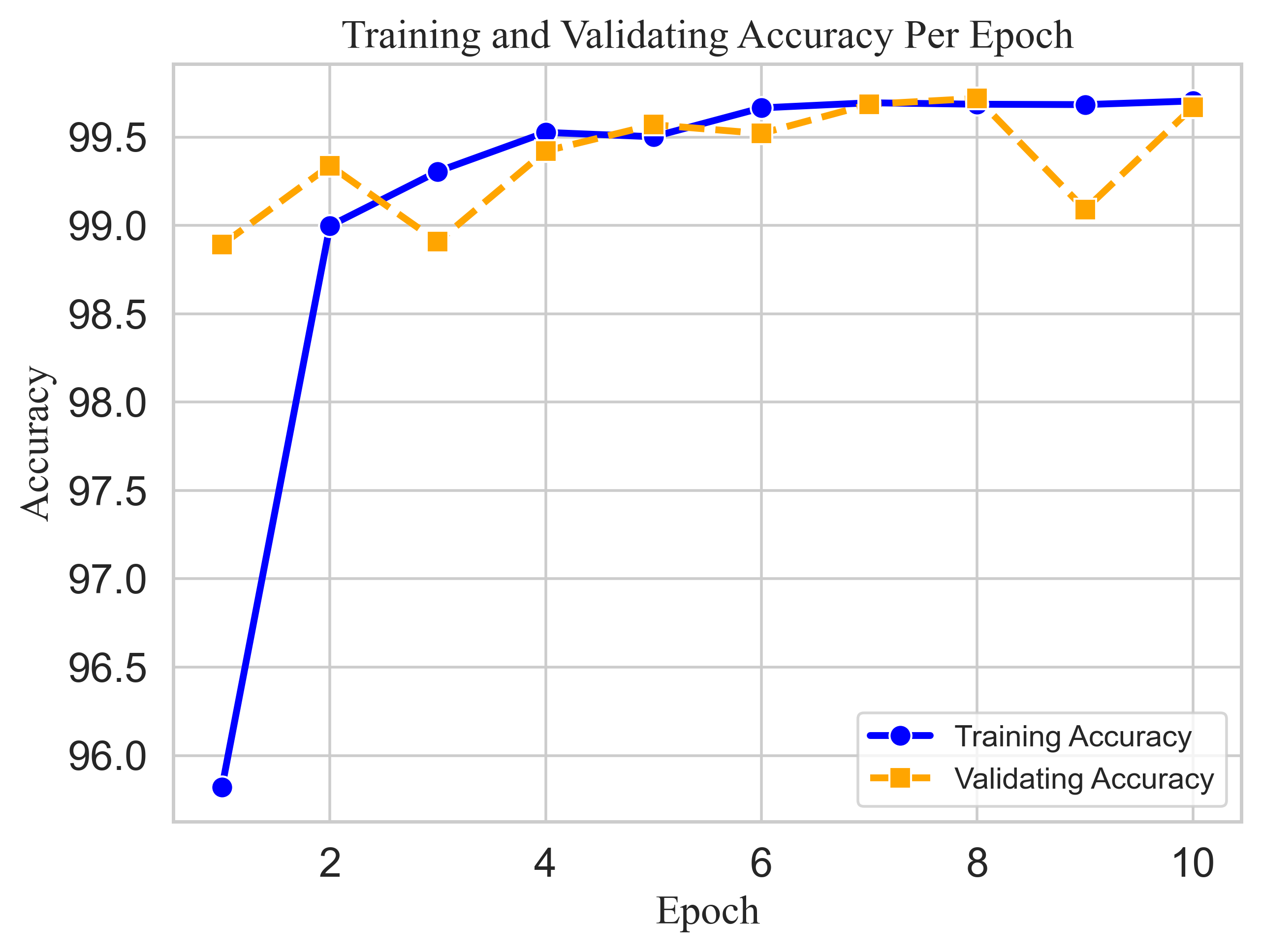}}
      \subfigure[]{\includegraphics[scale=0.5]{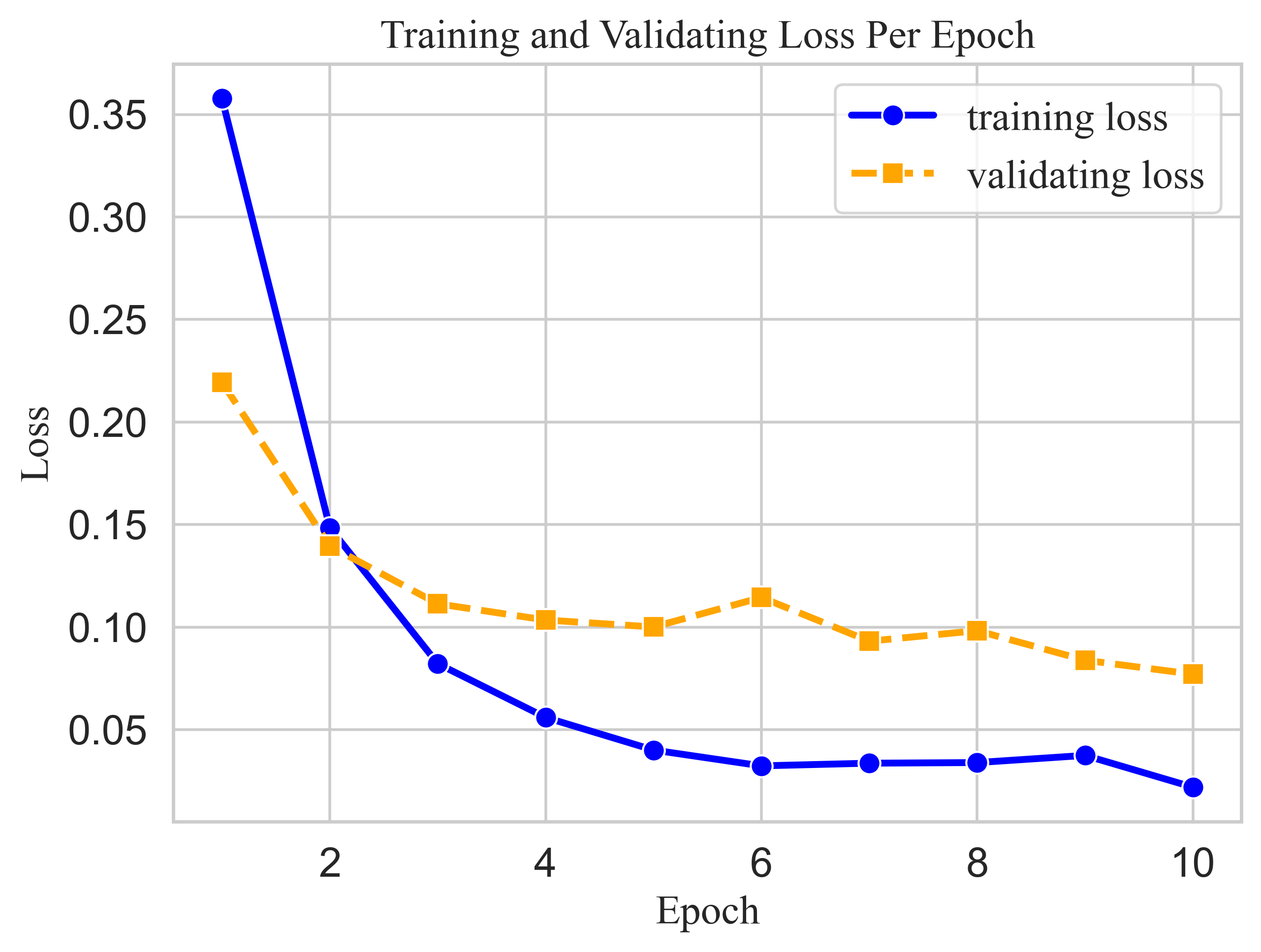}}
 \subfigure[]{\includegraphics[scale=0.5]{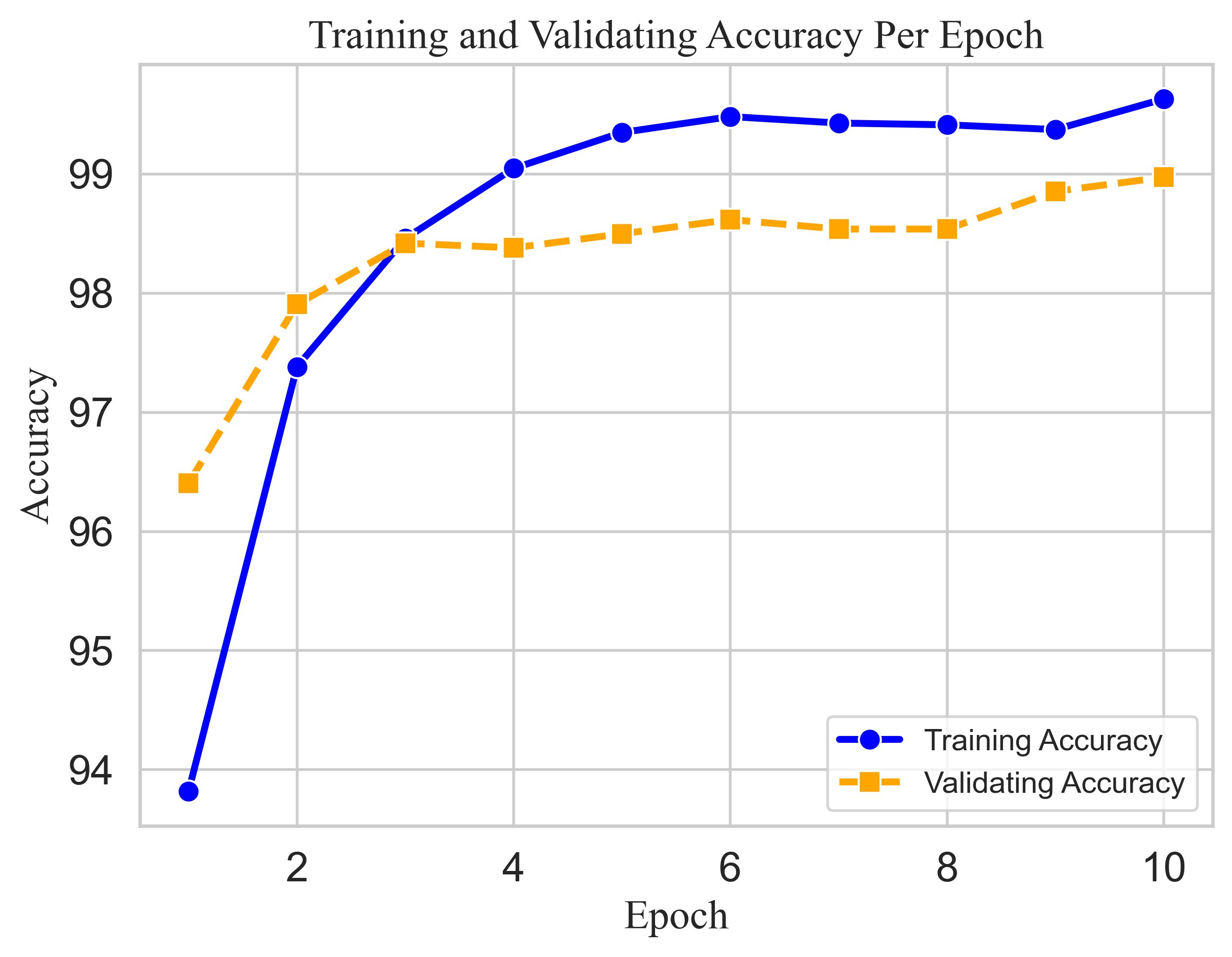}}
      \caption{The accuracy and loss curves in training and validating processes on datasets with and without augmentation. (a) Training and validating losses with augmentation. (b) Training and validating accuracies with augmentation. (c) Training and validating losses without augmentation. (d) Training and validating accuracies without augmentation. }
  \label{fig:fig3}
\end{figure*}

\section{Experiments}\label{s4}
In this section, we conduct experiments using the new dataset with and without data augmentation to validate the dataset and evaluate the performance of the DTViT. The experiment environment and parameters used in the training process are first given. Then, we introduce several evaluation metrics used in experiments. Further, we present the accuracy and losses for methods evaluated on the dataset. In addition, we apply classical CNN models to the dataset for comparison to depict the performance of our model. 

\subsection{Environments and parameters}
The experimental hardware setup is as follows: The CPU used is an AMD Ryzen 5975WX with 32 cores, and the GPU is an NVIDIA RTX 4090 equipped with 24 GB of memory. The operating system is Ubuntu 20.04. CUDA version 12.3 was utilized for computation. All experiments were conducted using Python version 3.8.18 and PyTorch version 2.3.0. 

Additionally, we utilized the AdamW optimizer \cite{loshchilov2017decoupled} with an initial learning rate of $2 \times 10^{-5}$ and a weight decay rate of 0.01. The batch size is set as 8 for the training data without data augmentation,  32 for the training data with data augmentation, 32 for the validating data,  and 4 for the testing data. The model was trained over ten epochs. We employed the cross-entropy function as the loss function to optimize the model's performance.
\setlength{\tabcolsep}{8pt} 
\renewcommand{\arraystretch}{1.3} 
\begin{table*}[h]
\centering
\caption{Performance comparison of DTViT with and without data augmentation on the testing dataset.}
\begin{tabular}{ccccccc}
\hline
    \textbf{Classifier} & \textbf{DA} & \textbf{Accuracy} & \textbf{Precision} & \textbf{Recall} & \textbf{F1 Score} & \textbf{Specificity} \\
\hline
Classifier 1 & Yes & 1 & 1 & 1 & 1 & 1 \\
Classifier 1 & No & 0.996 & 0.991 & 1 & 0.995 & 0.995 \\
Classifier 2 & Yes & 0.996 & 0.998 & 0.997 & 0.997 & 1 \\
Classifier 2 & No & 0.992 & 0.994 & 0.996 & 0.995 & 0.995 \\
\hline
\end{tabular}\label{table2}
\end{table*}

\subsection{Evaluation indices}\label{2.3.3}
We have conducted various experiments and adopted multiple models to evaluate the performance of DTViT. Firstly, the correct classification of images, i.e., the accuracy, is the criterion of the model performance, which is presented as 
\begin{equation}
   \text{Accuracy}=\frac{TP+TN}{TP+TN+FP+FN},
\end{equation}where $TP$ represents true positive instances; $TN$ represents true negative instances; $FP$ represents false positive instances; $FN$ represents false negative instances. The precision metric represents the proportion of correctly predicted positive instances out of all the instances predicted as positive in a class. Precision is determined using the following equation:
\begin{equation}
    \text{Precision}=\frac{TP}{TP+FP}.
\end{equation}The recall metric quantifies the proportion of positive instances that are correctly identified, and the corresponding  formula to calculate recall is presented as\begin{equation}
    \text{Recall}=\frac{TP}{TP+TN}.
\end{equation}In addition, the F1-score metric is the harmonic mean of precision and recall. It is computed as \begin{equation}
    \text{F1}=\frac{2*(\text{Precision}*\text{Recall})}{\text{Precision}+\text{Recall}}.
\end{equation}In medical fields, specificity is a good evaluation index to guarantee that individuals who do not have the disease are not wrongly diagnosed, thereby avoiding unnecessary treatment and anxiety. The calculation of specificity is presented as 
\begin{equation}
    \text{Specificity}=\frac{TN+FP}{TN}.
\end{equation}We take these metrics to evaluate the performances of DTViT on the newly collected dataset with and without data augmentation.

\begin{figure*}[h]
  \centering
     \subfigure[]{\includegraphics[scale=0.32]{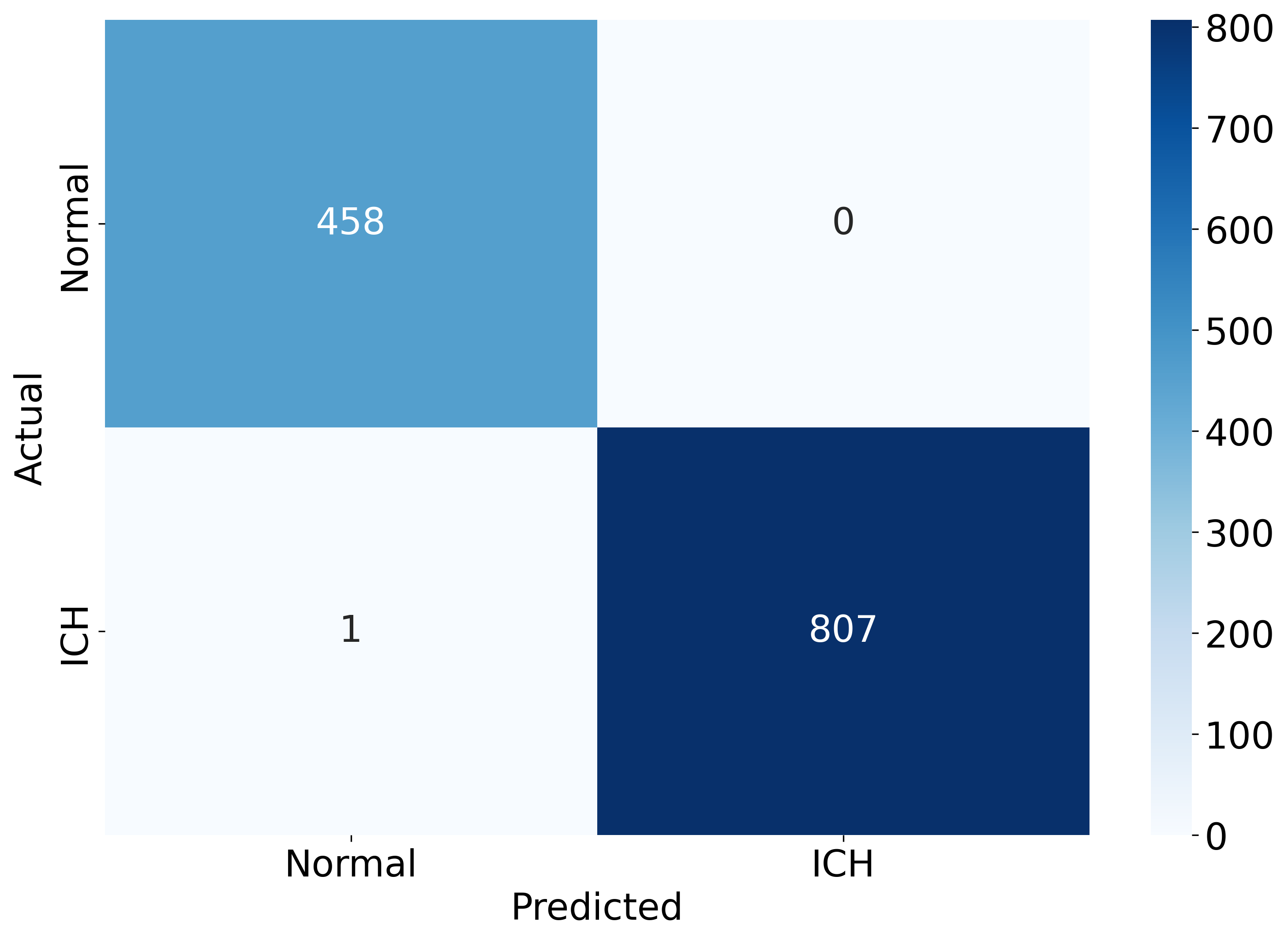}}
 \subfigure[]{\includegraphics[scale=0.32]{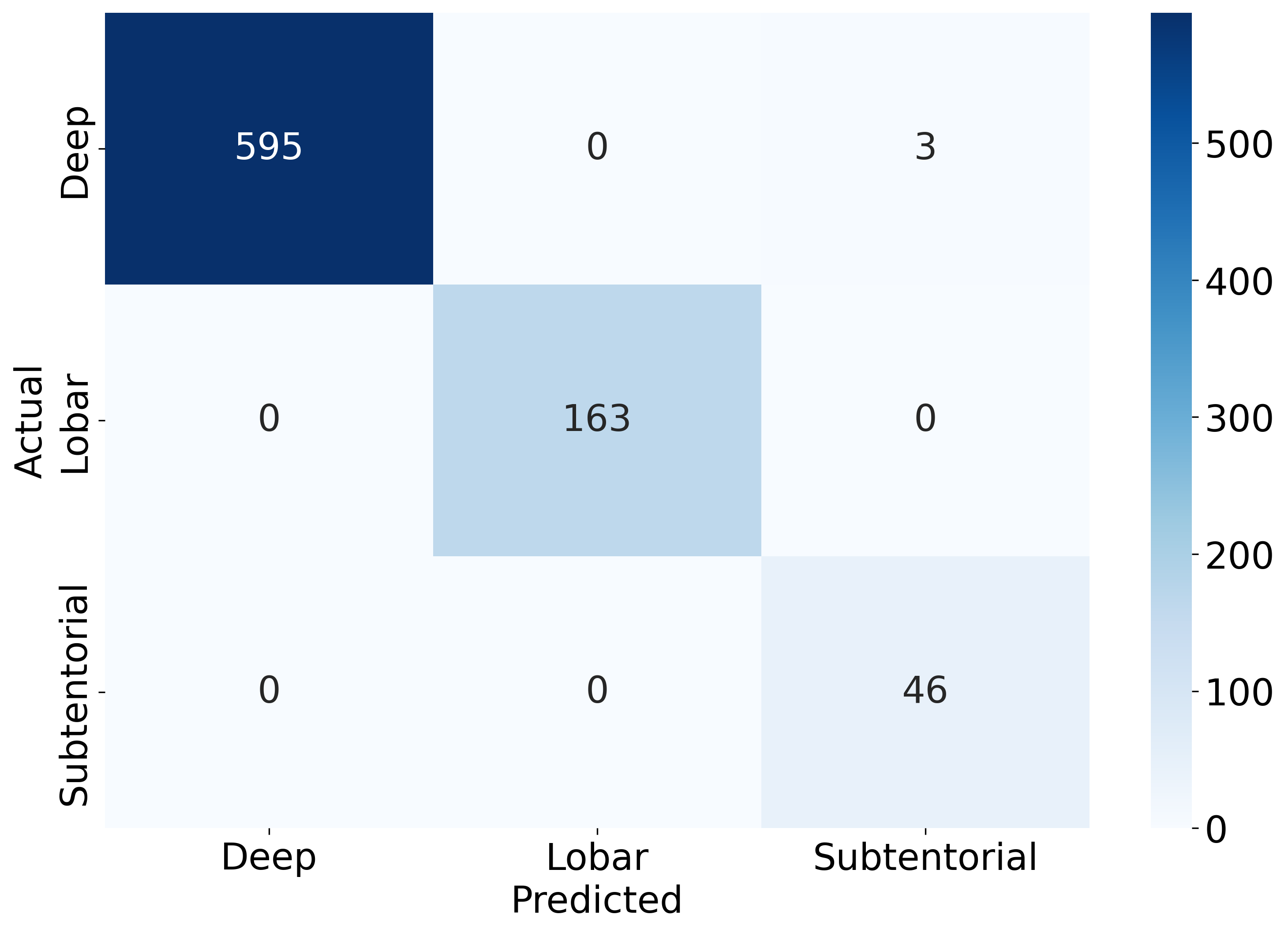}}
      \caption{Confusion matrixes of DTViT for two classification tasks on the testing dataset. (a) Confusion matrix of task 1 for normal and ICH classification. (b) Confusion matrix of Task 2 for three types of ICH classification.}
  \label{fig:fig4}
\end{figure*}

\subsection{DTViT performances}
 Figure \ref{fig:fig3} illustrates the evolution of losses and accuracies during training and validating processes, both with and without data augmentation. As depicted in Fig. \ref{fig:fig3}(a),  training and validating losses decrease steadily throughout the process, eventually stabilizing at approximately 0.01. Correspondingly, both training and validating accuracies show steady increases, converging approximately to 0.997 by the end of the training period. By contrast, results from the non-augmented dataset display inconsistency: while training loss decreases continuously to a desirable value, the validating loss fails to converge to the same level, ending around 0.08 and indicating overfitting. Similarly, the validating accuracy does not reach the level of the training accuracy,  demonstrating that the model is already sufficiently trained.

Further, we evaluate the trained model on the test dataset to obtain the DTViT's performance on the real-world untrained data, which is composed of 1266 CT images, including 459 for normal images and 807 ICH images with Deep 595, Lobar 163, and Subtentorial 49 images. Confusion matrixes are shown in Fig. \ref{fig:fig4}, where Fig. \ref{fig:fig4}(a) shows the result of Task 1 for normal and ICH patients classification, and Fig. \ref{fig:fig4}(b) shows the results of Task 2 for three types of ICH classification. It can be seen that almost all positive and negative cases on the test dataset are classified correctly, except one normal image is misclassified as an ICH image. Similarly, the classification of ICH types also achieved excellent results with a testing accuracy of $0.996$. In detail, evaluation indices of precision, recall, F1 score, and specificity are displayed in Table \ref{table2}.

\begin{table*}[h]
\centering
\caption{Overall performance comparisons with existing models.}
\begin{tabular}{ccccccc}
\hline
 \textbf{Model} & \textbf{Loss} &\textbf{Accuracy} & \textbf{Precision} & \textbf{Recall} & \textbf{F1 Score} & \textbf{Specificity} \\
\hline
\textbf{Proposed} & 0.0102& 0.9988 & 0.9991 & 0.9983 & 0.9987 & 1\\
ResNet18 \cite{he2016deep} &  0.0170& 0.9984 & 0.9989 & 0.9983 &0.9986  &0.9993 \\
AlexNet \cite{krizhevsky2012imagenet} & 0.0142 & 0.9976 & 0.9989 & 0.9983 &0.9986  &0.9994\\
SqueezeNet \cite{iandola2016squeezenet} & 0.1244  & 0.9775 &0.990& 0.9889 &0.9895 &0.9993 \\
DenseNet \cite{huang2017densely} & 0.0119 & 0.9976 & 1 & 0.9964 & 0.9981 &1 \\
\hline
\end{tabular}\label{table3}
\end{table*}

\subsection{Comparative experiments}
To better demonstrate the performance of DTViT, 
we conducted comparative experiments using classical CNN models, including ResNet18 \cite{he2016deep}, AlexNet \cite{krizhevsky2012imagenet}, SqueezeNet \cite{iandola2016squeezenet}, and DenseNet \cite{huang2017densely} on our augmented dataset. In the experiments, encoders of these models are adopted, and the decoder is set as the MLP classifier as DTViT. Additionally, these models are trained for ten epochs based on the transfer learning that pre-trained models are loaded for fine-tuning. The optimizer and essential parameters are all identical.

Comparative results are shown in Table \ref{table3}. It can be seen that all models have excellent accuracies of over $97\%$ on the augmented dataset. Specifically, the proposed model, DTViT, achieves the lowest error of $0.0102$ and the highest accuracy of $0.9988$ over these models. In addition, the DTViT also ranks first on other evaluation metrics, except its precision is slightly lower than DenseNet's.

\section{Discussion}\label{s5}
In this paper, we first collect CT images from intracerebral hemorrhage (ICH) patients and normal people, which are sourced from real-world patient data from Yulin First Hospital. Furthermore, medical specialists put effort into data filtering and labeling of three types of ICH  
haemorrhage.

Based on the new dataset, we propose the dual-task vision transformer (DTViT) model based on the vision transformer for dual-task classification. The proposed model is composed of an encoder to extract information from CT images and two decoders for different classification tasks, i.e., classification of normal and ICH images and classification of three types of ICH based on the location of the hematoma. The proposed DTViT achieves 99.7\% of the training accuracy and 99.88\% of the testing accuracy on the augmented dataset. We also compared the proposed model with multiple state-of-the-art models, i.e., Resnet18, SqueezeNet, and Alexnet, and the results show that our model achieves the best performance over these models. 

There is some previous research focusing on ICH classification or detection \cite{li2024classification,cortes2023deep,chen2022deep,raposo2023causal}. For example,  \cite{li2024classification} proposes an ICH classification and localization method using a neural network model, achieving an accuracy of 97.4\%, while the input signals are microwave signals and the hardware requirements are relatively high. Also, a CT-image-based deep learning method is proposed in \cite{cortes2023deep}, which is based on the EfficientDet and achieves an accuracy of 92.7\%. Similar work using computer vision methods for ICH detection and classification includes \cite{phaphuangwittayakul2022optimal,altuve2022intracerebral}, where  \cite{phaphuangwittayakul2022optimal} achieves ICH classification using a CNN-based architecture called EfficientNet; \cite{altuve2022intracerebral} uses ResNet-18 for ICH classification with the accuracy  95.93\%.

However, to the best of our knowledge, there is neither such a dataset describing the location of hematoma of ICH CT images nor such classification model simultaneously determining whether a CT image is of a cerebral hemorrhage or normal and classifies the three types of cerebral hemorrhage.

The primary aim of creating the Intracerebral Hemorrhage (ICH) CT image dataset and developing the classification model is to harness state-of-the-art computer vision technology to assist physicians in diagnosing and treating patients with cerebral hemorrhage. Cerebral hemorrhage is a severe, acute medical condition affecting approximately two million individuals annually, often associated with higher mortality and morbidity and limited treatment options. Early diagnosis and tailored treatment strategies based on the hemorrhage location are crucial for patient outcomes, as different hemorrhage sites require varied treatment approaches. Therefore, the classifier developed in this study, which categorizes hemorrhage based on its location, is clinically significant as it aids healthcare professionals in quickly and accurately determining treatment plans.

\textbf{Limitations:} This study faces several limitations. First, the dataset, derived from real clinical data, is challenging to obtain and inherently imbalanced, which may limit the model's accuracy and generalizability. Future efforts will focus on expanding and balancing the dataset through continued data collection. Second, the current data augmentation technique involves mere replication of the dataset. Future improvements will explore the use of generative models, such as diffusion models, for data enhancement.

\textbf{Future Directions:} In subsequent work, we aim to develop a multimodal diagnostic dataset for cerebral hemorrhage, integrating clinical data such as blood pressure, lipid profiles, and bodily element levels to enhance diagnostic accuracy. Furthermore, we plan to create a cerebral hemorrhage classification and diagnosis model based on Artificial Intelligence-Generated Content (AIGC), which will improve diagnostic efficiency for physicians.

\section{Conclusion}\label{s6}
In this paper, we have constructed a dataset for Intracerebral Hemorrhage (ICH) based on real-world clinical data. The dataset comprises CT images that have been initially processed and manually labeled by medical experts into categories of normal and ICH images. Moreover, the ICH images have been categorized into three types, Subcortical and Lobar, according to the hemorrhage's location.

Additionally, we have introduced a dual-task vision Transformer (DTViT) aiming at the classification of Intracerebral hemorrhage. This innovative neural network model incorporates an encoder, which utilizes the cutting-edge vision Transformer architecture, and two decoders that are designed to classify images as ICH or normal and to determine the hemorrhage type. Our experiments have demonstrated that the DTViT has achieved a remarkable accuracy rate of $99.88\%$ on the test data. We have also compared the DTViT with classical models such as Resnet18, SqueezeNet, and Alexnet.

To our knowledge, this research is the first to have developed both a specialized dataset and a neural network model tailored for hemorrhage location classification in a clinical context. This contribution holds substantial potential for enhancing clinical diagnosis and treatment planning.

\section*{Acknowledgments}
This work was supported by Yulin Science and Technology Planning Project under YF-2021-34.
\bibliographystyle{unsrt}  
\bibliography{main}  
\end{document}